# Efficacy of AI RAG Tools for Complex Information Extraction and Data Annotation Tasks: A Case Study Using Banks' Public Disclosures[1]

Nicholas Botti[2], Flora Haberkorn, Charlotte Hoopes, Shaun Khan[3]

Abstract

We utilize a within-subjects design with randomized task assignments to understand the effectiveness of using an AI retrieval augmented generation (RAG) tool to assist analysts with an information extraction and data annotation task. We replicate an existing, challenging real-world annotation task with complex multi-part criteria on a set of thousands of pages of public disclosure documents from global systemically important banks (GSIBs) with heterogeneous and incomplete information content. We test two treatment conditions. First, a "naive" AI use condition in which annotators use only the tool and must accept the first answer they are given. And second, an "interactive" AI treatment condition where annotators use the tool interactively, and use their judgement to follow-up with additional information if necessary. Compared to the human-only baseline, the use of the AI tool accelerated task execution by up to a factor of 10 and enhanced task accuracy, particularly in the interactive condition. We find that when extrapolated to the full task, these methods could save up to 268 hours compared to the human-only approach. Additionally, our findings suggest that annotator skill, not just with the subject matter domain, but also with AI tools, is a factor in both the accuracy and speed of task performance.

[1] The views expressed in this paper are solely those of the authors, and do not represent the views of the Board of Governors of the Federal Reserve System or the US Department of the Treasury.
We thank Jonathan Tong for research assistance. We thank Daniel Beltran for his mentorship and comments, Ricardo Correa for comments and suggestions. We also thank our discussant Roeland Beerten, and participants at the 4th BIS-IFC Workshop on Data Science in Central Banking for feedback and comments. All errors are our own.
We thank the authors of the original study referenced for generously allowing us to use their dataset and for their assistance with various aspects of this analysis: Cord Barnes, Daniel Beltran, Kellen Lynch, Monica Sanz, and Pinar Uysal.
[2] Corresponding Author Email: nicholas.botti@frb.gov
[3] Nicholas Botti, Flora Haberkorn, and Charlotte Hoopes are affiliated with the Board of Governors of the Federal Reserve System. Shaun Khan is affiliated with the US Department of Treasury but his work was performed while affiliated with the Board of Governors of the Federal Reserve System.

1. Introduction

Recent advancements in artificial intelligence (AI) tools for document analysis and information extraction, particularly the development of large language models (LLMs) and retrieval augmented generation systems (RAG) have generated intense interest in social science research and among policy analysis. The IMF's 2024 Annual Report cited how AI will transform the work of many industries (IMF 2024). Therefore, it is imperative for us to understand both when and how to use these tools effectively and safely, especially in the realm of economics, finance, and policy analysis.

One particular use case that has drawn considerable interest in policy analysis, is the ability to turn unstructured comprehensive and widely available text into structured datasets suitable for use for further assessment. Using unstructured text data to create a clean structured dataset suitable for financial and economic analysis has historically been a time-consuming process that starts with manual data collection. While some data vendors already use certain AI tools to extract data from unstructured text documents and make these datasets commercially available, they are often produced with a lag. And this kind of vendor data is not necessarily available for many niche topics. Our study helps bridge this gap by exploring the effectiveness of AI tools for extracting useful information quickly from unstructured data sources, such as public reports and investor disclosure statements by banks.

There is often a need to be able to access information on economic and financial conditions quickly. The example of Silicon Valley Bank (SVB) is instructive. It has been called the first "digital bank run," having taken place over a short period of time, and requiring investors and policymakers to act quickly based on incomplete information (Tett 2023).

Rather than studying a complex data pipeline, we seek to understand how financial analysts can effectively extract this type of information on an ad-hoc basis in a real-world task setting. This is consistent with available information that suggests that thus far the most common way of using these AI tools is as a personal productivity assistant (Maslej, et al. 2025).

Therefore, we set out to study how effective human annotators and analysts are at this type of task, utilizing a within-subjects design to model the use of AI tools as a treatment effect.

We investigate the accuracy of this approach, the magnitude of time-savings, as well as determining the effect of other factors on these outcomes.

## 2. Literature Review

### AI as an Assistant and Productivity

Numerous studies have been published focusing on AI and productivity, such as the study done on how introducing a generative AI tool benefited junior software engineers compared to their seniors (Jaffe, et al. 2024) as well as a review done by the BIS that found using LLM's can boost productivity of engineers (Gambacorta, et al. 2024). Investigation into knowledge workers productivity has shown only half the time savings gained as software engineers (Bick, et al. 2025), but doesn't review a variety of factors such as how the worker's background and AI knowledge is a factor in how effective they are in using the tool.

### Measuring the Performance of AI Models on Tasks

The literature in artificial intelligence and machine learning is dedicated to studying the performance of LLMs on benchmarks and has continued to struggle to capture the speed of development and domain-specific applicability. ARC (Chollet 2019) and DROP (Dua, et al. 2019) are two of the premier assessment tools used to evaluate the ability for an LLM to gather important information from documents in a manner that emulates research assistance. RAG specific benchmarks such as TAT-QA (Zhu, et al. 2021), cover the effectiveness of finance information retrieval while benchmarking other domains. However, these benchmarks and others often fail to capture the complexities of real-world tasks, particularly in the economics and financial policy domains. Even with the introduction of agents, few benchmarks have emerged looking at making economic-based choices such as EconEvals (Fish, et al. 2025). However, such benchmarks often also focus on automated analysis, rather than the effects of pairing a human analyst with these tools, leaving them limited from a qualitative scope.

      In our study, we investigate a real-world information extraction and data annotation task previously completed by human analysts. By replicating an existing, completed task, we can

create a robust evaluation of the impact of introducing these AI tools on this specific workflow. Additionally, we structure our evaluation as a within-subjects design, allowing us to understand the effect of introducing use of the AI tool as a treatment effect compared with a control human-only approach, all while controlling effectively for per-annotator skill.

## 3. Methodology

3.1 Case study using banks' public disclosures.

We utilized the annotated dataset compiled by Beltran et al. (2023), which was created through a data collection process conducted by human annotators in 2023 to extract diverse information from bank documents. The goal of the collection was to extract information from the public and investor disclosure documents of global systemically important banks (GSIBs), and create an annotated dataset of GSIB policies, commitments, and progress on certain key metrics. This unified dataset across banks would allow researchers to make cross-bank comparisons and better understand industry trends over time. This dataset was compiled using a human only approach, so it offers a natural benchmark to compare to our RAG approach.

Human annotators examined thousands of pages of documentation per bank, across annual reports, non-financial disclosures, and specialized bank reports for each financial institution. The targeted information ranged from banks' broad investment goals to banks exact dollar exposure to certain sectors. The complete dataset included 150 questions for each of the 29 GSIBs, a mix of true/false questions, categorical assessments of policies and activities, and exact figure questions.

This task makes for an excellent case-study in assessing the impact of exposing analysts to AI tools to help complete their work. It is a complex, real-world task in the financial analysis domain, not a multiple choice question benchmark. The questions often required combining information from multiple different sources within the documents to produce a final answer. Additionally, because of heterogeneity in how banks disclose information, we can assess the efficacy of the AI-assisted approach under incomplete information. Because the documents do not always contain complete information to answer all the questions for every single bank, we were able to assess whether AI reacted appropriately in scenarios when information wasn't available by correctly refusing to answer.

## 3.2 Study Design

This study employed a mixed experimental design to evaluate the impact of an Artificial Intelligence (AI) Retrieval-Augmented Generation (RAG) tool on the performance of human annotators performing a complex information extraction task using public disclosure documents from Global Systemically Important Banks (GSIBs). The core experiment utilized a within-subjects design, where specific questions within each bank annotation task were randomly assigned to either a 'Control' condition or an 'AI' (naive use) condition for each participant. This allowed for comparing the performance under each condition while controlling for individual annotator differences.

A subsequent repeated-measures phase was conducted later, where participants revisited the questions previously assigned to the 'AI' condition, this time operating under an 'AI+' (interactive use) condition. This allowed for a direct comparison of naive versus interactive AI tool usage on the same set of questions while controlling for recall effects.

## 3.3 Participants

Three professional financial analysts participated in the study. Participants were recruited via a convenience sample of volunteers within the researchers' organization and completed the experimental tasks during their regular work hours. Prior to the experiment, participants completed a multiple-choice quiz designed to assess their baseline familiarity with both the subject matter (GSIB disclosures and related finance topics) and AI tools. These baseline skill assessments are used descriptively and for subsequent analysis exploring potential interactions between skill profiles and condition effects.

## 3.3 Materials

**Bank Documents:** The materials consisted of comprehensive sets of public disclosure documents retrieved from bank websites and formatted as pdf files (e.g., annual reports, non-financial reports, regulatory filings) for a sample of nine European GSIBs. These document sets were identical to those compiled and then used in the original analysis by Beltran et al. (2023) and were provided by the original research team, ensuring consistency and relevance.

**Annotation Questions:** A standardized set of 30 questions was used for all banks and participants. These questions were selected in consultation with the original Beltran et al. (2023) research team to be representative of the full 150-question set used in their study. The selection aimed for diversity across question complexity (high: 9, medium: 11, low: 6), required answer format (categorical: 9, Yes/No: 10, exact figure/value: 7), and thematic category (risk management: 11, financed activity: 10, oversight rules: 3, external project ratings: 2). Each question included detailed instructions and definitions identical to those used by Beltran et al. (2023).

## 3.4 Experimental Conditions

Participants completed annotation questions under three distinct conditions:

1. **Control Condition:** Participants answered assigned questions using only the provided PDF documents and a standard PDF reader. They were permitted to use the reader's built-in text search ("find") functionality but were explicitly prohibited from using any other external tools, web searches, or AI assistance. The focus was on manual information retrieval and synthesis based solely on the source documents.
2. **AI (Naive) Condition:** Participants answered assigned questions using a provided AI RAG chat tool pre-loaded with the relevant bank's document collection. Participants were instructed to formulate a single prompt based on the annotation question and submit it to the tool. They were explicitly instructed to record their final answer based *only* on the AI tool's initial response, without consulting the original PDF documents. This condition aimed to simulate a naive or high-trust approach to using the AI assistant.

3. **AI+ (Interactive) Condition:** This condition was implemented in the follow-up phase for questions previously tackled under the 'AI' condition. Participants initiated the process identically to the 'AI' condition (formulate single prompt, review AI response). However, they were then given discretion: based on their judgment of the initial AI response, they could choose to (a) ask additional follow-up questions within the AI chat tool, (b) consult the original PDF documents, or (c) do neither or both, before recording their final answer. This condition aimed to simulate a more interactive, discerning, and potentially iterative approach to using the AI assistant.

AI RAG Tool Specifications: The same AI RAG tool configuration was used for both the 'AI' and 'AI+' conditions. It was implemented using OpenWebUI (version 0.3.2) as the interface. The backend utilized Claude 3.5 Sonnet (model version 20240622) via Amazon Bedrock as the Large Language Model (LLM) with default sampling parameters. Vector embeddings were generated using Amazon Titan Text Embedding v2 via Amazon Bedrock. For the RAG process, documents were chunked using a size of 2000 characters with a 200-character overlap. The retrieval mechanism used a "top_k" setting of 20, meaning the 20 most relevant chunks were retrieved to inform the LLM's response. This setup was chosen to represent a commonly accessible, though not state-of-the-art, RAG implementation suitable for typical researcher workflows.

## 3.5 Procedure

Participants received initial training covering the task background, the experimental procedures for each condition, instructions on using the AI RAG tool interface, and guidelines for self-reporting time. The detailed instructions embedded within each annotation question served as the primary guide for content interpretation.

For each assigned bank, the 30 questions were presented in the same fixed order. Prior to the task, each question was randomly assigned (using a computer script with a random number generator, resulting in an approximately 50/50 split) to either the 'Control' or 'AI' condition for that specific annotator-bank instance. Randomization was not stratified by question characteristics.

Participants performed the tasks at their own discretion regarding location (e.g., office, remote). For each question, participants self-reported their start and end time (wall clock time) to the nearest minute. Durations under one minute were recorded as 0.5 minutes, based on observational validation for a subset of tasks.

The initial experiment (Control vs. AI) was completed first. The follow-up experiment (AI+) was conducted after a minimum delay of two weeks for all participants (up to three months for two participants) to mitigate potential recall effects from the initial phase.

Bank assignments were as follows: Annotator A completed tasks for six banks, Annotator B for three banks, and Annotator C for four banks. Annotator C's assignments included two banks also completed by Annotator A and two banks also completed by Annotator B, creating deliberate overlap to enable the assessment of inter-annotator agreement.

3.6 Outcome Measures

**Time Efficiency:** Measured as the self-reported time taken per question (in minutes) for each condition.

**Accuracy/Agreement:** The primary measure of answer quality. Annotator responses for all conditions were compared against the final, vetted answers established by the Beltran et al. (2023) research team. Although the realistic nature of this task implies that there are no known true values, given the thoroughness of the original research team's work, we consider comparison against these values as an acceptable proxy for task accuracy. A member of the original team performed this agreement assessment, consulting with other original team members for complex cases. For each answer, agreement was categorized as "exact match," "substantially similar," "unclear," or "incorrect." For the main analysis, "exact match" and "substantially similar" were coded as Agreement = True (representing acceptable answers based on the original team's criteria), while "incorrect" was coded as Agreement = False. Answers rated "unclear" (20 answers, roughly 4% of total answers) were excluded from the accuracy analysis. Robustness checks that did not count "substantially similar" as agreement yielded broadly similar results.

**AI+ Interaction Metrics:** For the AI+ condition only, participants recorded (a) the number of distinct follow-up messages sent to the AI tool after the initial prompt, and (b) whether they chose to consult the original PDF documents (Yes/No) before finalizing their answer.

## 3.7 Ethical Considerations

Informed consent was obtained orally from all participants prior to their involvement in the study. Participants were informed about the study's purpose, procedures, data handling, and the voluntary nature of participation. Data collected, including annotator responses, timing, and quiz scores, were de-identified before the main statistical analysis to protect participant confidentiality.

## 4. Results

### 4.1 Participant Characteristics

Prior to the study, participants were evaluated on a short assessment. The assessment consisted of two parts. The first portion was devoted to their skill and understanding of AI - assessed by questions about prompt techniques, characteristics of different models, understanding of the fundamentals of RAG, knowledge about hallucinations, etc. The second portion measured their understanding of the subject matter domain of the documents - regulatory frameworks, bank policies, balance sheet considerations, etc.

Participant Skill by Annotator

Table 1

| Participant | Subject Matter Skill (%) | AI Skill (%) |
|---|---|---|
| A | 68.8 | 52.2 |
| B | 68.8 | 82.6 |
| C | 87.5 | 21.7 |

As shown in Table 1 all three participants had generally high scores for their understanding of the subject matter domain, consistent with their roles, but their scores on the AI portion of the assessment varied much more, consistent with anecdotal questions we asked participants about their experience with AI tools. Participant B and C provide the largest contrast between their skill sets, while participant A is more in the middle.

4.2 Question Characteristics

The questions used in the task are generally well balanced, and are a representative subset of the full 150 question dataset produced by Beltran et al. 2023. Tables 2, 3, and 4 below outline the makeup of questions tested by annotators over the course of different treatments (a total of 494 question-answer pairs).

Question Formats

Table 2

| Format | Count | Percentage |
|---|---|---|
| Yes/No | 192 | 38.9 |
| Categorical (cat) | 178 | 36.0 |
| Numerical FIgure | 124 | 25.1 |

Question Complexity

Table 3

| Complexity | Count | Percentage |
|---|---|---|
| Low | 114 | 23.1 |
| Medium | 209 | 42.3 |
| High | 171 | 34.6 |

Question Categories

Table 4

| Category | Count | Percentage |
|---|---|---|
| Risk Management | 204 | 40.9 |
| Financing Activities | 199 | 40.3 |
| Oversight | 57 | 11.5 |
| External Organizations | 26 | 7.3 |

## 4.3 Time Efficiency

Time to Complete One Question, in Minutes

Table 5

| Condition | Obs | Mean | SD | Median | Min | Max |
|---|---|---|---|---|---|---|
| Control | 153 | 4.0 | 2.7 | 4.0 | 0.5 | 15 |
| AI | 168 | 0.3 | 0.3 | 0.5 | 0.5 | 1 |
| AI+ | 173 | 1.2 | 1.2 | 1.0 | 0.5 | 6 |

AI assistance reduced task completion time by 90% in the naive AI condition and 70% in the interactive AI+ condition, with minimal variance. It should be noted that this time to answer the question represents a first-pass or "best effort" attempt to answer the question. While we do not have exact figures, the original research team reports that they spent on average significantly longer than this per question before arriving at their final results - such as by incorporating multiple levels of review and discussing some questions as a group.

If these numbers generalize to the full task (150 questions per bank, 29 banks), we would estimate the following time savings:

Time Saving for Complete Task (Estimated)

Table 6

| Condition | Mean time per Question (mins) | Total Time (Estimated Hours) | Total Savings vs Control (Estimated Hours) |
|---|---|---|---|
| Control | 4.0 | 290 | - |
| AI | 0.3 | 21.75 | 268.25 |
| AI+ | 1.2 | 87 | 203 |

Time Efficiency by Annotator

Table 7

| Participant | Obs | Mean | SD | Median | Min | Max |
|---|---|---|---|---|---|---|
| A | 227 | 2.2 | 2.8 | 0.5 | 0.5 | 15 |
| B | 118 | 2.0 | 2.0 | 1.0 | 0.5 | 12 |
| C | 149 | 1.2 | 1.0 | 1.0 | 0.5 | 5 |

[1] Time reported in Minutes.

Average Time Taken per Task by Participant by Condition

Table 8

| Participant | Obs | Control | AI | AI+ |
|---|---|---|---|---|
| A | 227 | 5.5 | 0.5 | 1.0 |
| B | 118 | 3.9 | 0.6 | 1.8 |
| C | 149 | 1.9 | 0.7 | 1.2 |

[1] Time reported in Minutes. Average time calculated with simple mean.

Participant C was the fastest overall, with the lowest mean and median times, possibly due to stronger subject matter knowledge (87.5%) helping streamline the annotation process. Note the high variation in task time, ranging from 0.5 to almost 15 minutes depending on the question. Also note the unusually high amount of time taken by Participant B during the interactive AI condition, perhaps indicating their high level of experience with AI tools allowed them to more effectively take advantage of the capability.

## 4.4 Accuracy (Agreement)

Agreement Rates by Condition

Table 9

| Condition | Total Responses | Agreements | Agreement Rate (%) |
|---|---|---|---|
| Control | 153 | 81 | 52.9 |
| AI | 168 | 99 | 58.9 |
| AI+ | 173 | 114 | 65.9 |

Descriptively, accuracy increased somewhat with AI use. The highest agreement rate occurred in the AI+ condition, suggesting that an interactive approach allows annotators to better assess and refine AI outputs.

Agreement by Question Type and Condition

Table 10

| Format | Control(%) | AI (%) | AI+(%) | AI+ vs Control (pp) |
|---|---|---|---|---|
| Yes/No | 62.5 | 76.6 | 84.4 | +21.9 |
| Categorical | 29.2 | 35.9 | 40.9 | +11.7 |
| Numerical | 65.9 | 67.5 | 76.7 | +10.8 |

Unsurprisingly, annotators performed best on the True/False questions, and worst on the categorical questions, which often required tricky subjective judgement in how to assess whether the information met complex criteria. We see the greatest increase in performance with the AI+ method on these simpler yes/no questions.

Agreement by Complexity and Condition

Table 11

| Complexity | Control(%) | AI(%) | AI+ (%) | AI+ vs Control (pp) |
|---|---|---|---|---|
| Low | 66.7 | 76.3 | 90.0 | +23.3 |
| Medium | 61.3 | 69.9 | 73.0 | +11.7 |
| High | 34.5 | 33.3 | 40.7 | +6.2 |

Consistent with the findings from the question type, using the AI tool had the strongest positive effect on lower complexity questions. For higher complexity questions, it still improved accuracy somewhat, but much less so. Note the large variation in performance, consistent across conditions, between the low/high complexity questions, validating the effectiveness of the complexity assessments

Agreement by Participant

Table 12

| Condition | Total Responses | Agreements | Agreement Rate (%) |
|---|---|---|---|
| A | 227 | 123 | 54.2 |
| B | 118 | 72 | 61.0 |
| C | 149 | 99 | 66.4 |

Participants' overall accuracy scores were generally consistent with their evaluations of subject matter expertise. But this masks strikingly high variation when breaking down by experimental condition.

Agreement by Participant and Condition

Table 13

| Participant | Control(%) | AI(%) | AI+ (%) | AI+ vs Control (pp) |
|---|---|---|---|---|
| A | 48.6 | 53.2 | 60.3 | +11.7 |
| B | 33.3 | 68.3 | 75.0 | +41.7 |
| C | 72.9 | 60.0 | 66.7 | -6.2 |

Annotator B, who had by far the highest experience working with AI tools, was able to more than double their chance of correctly answering the question when using the AI tool, despite performing the worst in the control condition. Meanwhile Annotator C, who had the highest subject matter expertise but very little experience with AI tools, actually performed somewhat worse under the treatment conditions.

While the sample size here is clearly insufficient (three annotators) to generalize these findings, this is a very interesting result nonetheless that we believe should motivate future research.

5.  Statistical Analysis

5.1 Accuracy

We perform a logistic regression on the binary variable representing whether an answer (observation) correctly agrees with the findings of the original research team, per the agreement methodology discussed previously. We control for both the individual baseline of each annotator and the question complexity. We use the "Control" condition, annotator A, and medium complexity as the baseline. Dummy variables for the AI and AI+ conditions, annotators B and C, and complexity of high or low are included in the regression.

Our hypothesis for this experiment was that there would be a considerable increase in speed under the AI conditions, particularly the naive AI condition, but that there would be some loss of accuracy. We hypothesized that the key question would be how large that loss of accuracy would be, and under what conditions that loss of accuracy might be acceptable for certain tasks

We find that under the naive AI condition, accuracy increases, but is not statistically significant whereas under the interactive AI condition, accuracy increases significantly (p=0.014). High question complexity, and Annotator C both significantly decreased and increased accuracy (respectively).

$$\log(P(Agreement_i = 1)/(1 - P(Agreement_i = 1))) = \beta_0 + \beta_1 \cdot Condition\_AI_i$$
$$+ \beta_2 \cdot Condition\_AIplus_i + \beta_3 \cdot ParticipantID\_B_i + \beta_4 \cdot ParticipantID\_C_i$$
$$+ \beta_5 \cdot Complexity\_Low_i + \beta_6 \cdot Complexity\_High_i$$

Logistic Regression on Agreement_Boolean

Table 14

| Variable | Human vs AI vs AI+ n = 494 |
|---|---|
| Intercept | 0.2554 |
|  | (0.231) |
| Condtion_AI | 0.2557 |
|  | (0.243) |
| Condition_AI+ | 0.6010* |
|  | (0.246) |
| Participant_B | 0.2484 |
|  | (0.249) |
| Participant_C | 0.5655* |
|  | (0.236) |
| Complexity_Low | 0.5257 |
|  | (0.274) |
| Complexity_High | -1.3483*** |
|  | (0.221) |

[1] Signif. codes: '***' 0.001 '**' 0.01 '*' 0.05

Additionally, we also investigate whether checking the original documents or changing the number of messages to send to the AI affects the accuracy during the interactive (AI+) phase. Filtering to the AI+ (interactive) condition only, we include a dummy variable for whether the original documents were checked or not, as well as the number of messages sent to the AI tool before recording a final answer to the question.

Logistic Regression on Agreement_Boolean, Within AI+ Condition Only

Table 15

| Variable | AI+ (n = 173) |
|---|---|
| Intercept | 1.1349** |
|  | (0.05400) |
| **Doc_Check** | 0.0466 |
|  | (0.396) |
| ParticipantID_B | 0.9713 |
|  | (0.547) |
| ParticipantID_C | 0.2191 |
|  | (0.411) |
| Question_Complexity_low | 1.2258* |
|  | (0.597) |
| Question_Complexity_high | -1.2821*** |
|  | (0.385) |
| **Num_Messages** | -0.3787 |
|  | (0.269) |

[1] Signif. codes: '***' 0.001 '**' 0.01 '*' 0.05

We find that neither the number of messages sent and whether the analyst checks the original documents are statistically significant, which lends to multiple interpretations. This could indicate that these additional measures weren't helpful, while on the other hand this could be a selection effect. Thus, indicating that the analysts are correctly using their judgement to apply these measures only to harder questions instead of leveraging the AI for tasks they can complete confidently. We consider the latter the more straightforward and likely explanation, but further work could explore setting this as an explicit experimental condition. Analysts checked the original document 24% of the time for low complexity questions, but 38% and 39% of the time for medium and high complexity questions respectively, giving at least some indication that their judgement of when these additional steps were necessary was correct.

## 5.2 Time per Task

We then run a generalized linear mixed model (GLMM) to evaluate the effects of AI on time spent to answer a question. We include a gamma distribution and a log link to address the positive and right-skewed characteristics of our time data.

$$log(E[Time_{it}]) = \beta_0 + \beta_1 Condition_i + \beta_2 Question\_Complexity_i + \beta_3 ParticipantID\_B_i + \beta_4 ParticipantIDC_i + \beta_5 Complexity\_Med_i + \beta_6 Complexity\_High_i + u_i$$

Our results suggest that AI assistance significantly reduces the time required for annotation. Complexity also has a marginal effect, implying that more complex questions may demand more time. Additionally, we notice individual differences among annotators. The results show that annotator B spent significantly more time answering the questions. This could be attributed to the annotator's lower institutional knowledge on the subject or their higher skill with AI tools leading them to use slower but more advanced techniques.

GLMM Regression on Time_Minutes

Table 16

| Variable | Human vs AI vs AI+ (n = 494) |
| --- | --- |
| Intercept | 1.0629*** |
|  | (0.05400) |
| Condtion_AI | -1.65987*** |
|  | (0.05824) |
| Condition_AI+ | -0.96922*** |
|  | (0.05809) |
| Participant_B | 0.21772*** |
|  | (0.05907) |
| Participant_C | -0.03431 |
|  | (0.05513) |
| Complexity_Low | -0.12708* |
|  | (0.06003) |
| Complexity_High | 0.07441 |
|  | (0.05307) |

[1] Signif. codes: '***' 0.001 '**' 0.01 '*' 0.05

## 6. Conclusion

AI RAG tools show significant promise in helping analysts with extracting and categorizing information in lengthy datasets of financial documents in complex, real-world conditions. Even when using a relatively simple RAG architecture, we see no statistically significant decrease in task accuracy, and when used carefully and with judgement, we see a statistically significant increase in task accuracy. In both cases the time savings are significant, up to an order of magnitude. We find some evidence of the importance of training analysts on how to use these tools most effectively, and that these tools may struggle more with the most complex questions. Due to low sample size though, further research is needed into these areas.

Despite impressive improvements in accuracy for treated instances, it is important to note that we only measure here first-pass accuracy. Both in the control and AI treatment methods, accuracy still fails to breach 80% of what could be done with a more thorough team process involving multiple levels of review and verification. This underscores the importance of verifying critical information before making important decisions, with or without AI.